\documentclass[11pt]{article}
\usepackage[english]{babel}
\usepackage[utf8x]{inputenc}
\usepackage{amsmath,amsthm}
\usepackage{graphicx}
\usepackage[colorinlistoftodos]{todonotes}
\usepackage{listings}

\usetikzlibrary{shapes.geometric}
 
\title{\textbf{$\mathbf{\delta} \epsilon \alpha^*$\\ A PAC-Admissible Robotic Search Algorithm}}
\author{David Cox}
\date{July 2015}
\begin{document}

\maketitle
\begin{abstract}
In this paper, we introduce $\delta \epsilon \alpha^*$ -- a minimal coverage, real-time robotic search algorithm that yields a moderately aggressive search path with minimal backtracking. Search performance is bounded by a placing a combinatorial bound, $\epsilon$ and $\delta$, on the maximum deviation from the theoretical shortest path and the probability at which further deviations can occur.
Additionally, we formally define the notion of PAC-admissibility -- a relaxed admissibility criteria for algorithms, and show that PAC-admissible algorithms are better suited to robotic search situations than $\epsilon$-admissible or strict algorithms.  
\end{abstract}
\section{Introduction}
Robotic search presents several challenges that traditional search algorithms do not have to consider. Traditional search algorithms require the entire search space to be both bounded and mapped. In robotic search, it is often the case that at least one of these requirements cannot be met, and environmental data must be gathered through the process of exploration. Due to this limitation, robotic search necessitates that the search path be calculated in real-time. \\

While traditional search algorithms can be modified for use in real-time situations, they are inefficient due to their reliance on backtracking to ensure the shortest path is found. In a precomputed search, the process of backtracking does not add to the distance traveled, and thus the search produces the shortest path. In a real-time situation, backtracking is not free -- the robot must physically explore possible paths. Using traditional search algorithms in real-time situations would result in a path length that is greater that the sum of all considered paths due to backtracking. \\

Additionally, traditional search algorithms assume that the distance to the target is precisely known, while robotic search often is used in situations where the distance to the target is imprecise or possibly unknown if outside of a given range or line of sight. These uncertainties can be mitigated through exploration of the search space, however exploration is costly as it adds to the total distance traveled. It follows that a good real-time search algorithm would perform exploration in an intelligent manner to minimize unnecessary traversal of the space. \\

To address these challenges, we propose a real-time search algorithm that bounds the maximum search path within an error factor of $(1 +\epsilon)$ times the length of the shortest path with a probability of $(1-\delta)$. The PAC nature of this bound minimizes the amount of exploration and backtracking without producing an overly aggressive search. \\

\section{Background}
When considering the robotic search problem, it is unlikely that a direct, shortest path solution can be found given the degree of uncertainty present. Utilizing an approach with strict admissibility criteria would result in a significant amount of exploration and backtracking, making it impractical for a real-time search. \\

Better results could be had from a search that is $\epsilon$-admissible, as they avoid exploration and backtracking by favoring a promissory path, so long as the next node in the path is $\epsilon$-admissible. However, traditional $\epsilon$-admissible search algorithms are “broken” when applied to real-time situations -- the $\epsilon$ bound refers only to the final path and does not take into account the considered paths or the distance added due to backtracking. While $\epsilon$-admissible algorithms are less likely to backtrack than their stricter counterparts, they still backtrack to an unbounded degree and are therefore not ideal for robotic search.\\

\subsection{AlphA*}
AlphA* is an $\epsilon$-admissible search algorithm that has aggressive and conservative variations, both of which attempt to follow a promissory path as long as possible \cite{alpha*} Nodes in the search space are given priority as $f_\alpha=(1+w_\alpha (n))f(n)$ where the weight function is
$$
w_\alpha(n) =
\begin{cases}
\lambda, & \text{if }\alpha (n)\text{ is true} \\
\Lambda, & \text{otherwise}
\end{cases}
$$

and $\lambda$, $\Lambda$ are constants such that $-1 < \lambda \leq \Lambda$. Nodes are more likely to be expanded if their weight value is low, with the $\alpha$-perimeter $\alpha(n)$ being used to increase or decrease the likelihood of expansion.\\

AlphA* is presented with four possible $\alpha$-perimeters that represent cost and heuristic based aggressive and non-aggressive approaches to different usage situations e.g. a maze vs. small obstacles. One must be selected prior to the search and used for the entire duration.  \\

The $\alpha$-perimeters are defined as follows where $g(n)$ is the cost of a node, $h(n)$ the heuristic value, $\pi(n)$ the parent of $n$, and the most recently expanded node being $\hat{n}$. \\

Non-aggressive:
\begin{equation*}
\begin{split}
\alpha_g (n) \equiv g(\pi(n)) &\geq g(\hat{n})\\
\alpha_h (n) \equiv h(\pi(n)) &\leq h(\hat{n})
\end{split}
\end{equation*}

Aggressive:
\begin{equation*}
\begin{split}
\alpha_g (n') \equiv g(\pi(n)) &\geq \underset{n' \in \text{CLOSED}}{\max} g(n')\\
\alpha_h (n') \equiv h(\pi(n)) &\leq \underset{n' \in \text{CLOSED}}{\min} h(n')
\end{split}
\end{equation*}
To perform an efficient search with AlphA*, the correct $\alpha$-perimeter must be used. In an unknown search space, it is impossible to ensure proper pre-selection and thus ideal performance. Furthermore, in a real-time search, AlphA* is still subject to unbounded backtracking, rendering the search no longer $\epsilon$-admissible. 

\subsection{PAC Learning}
Probably Approximately Correct learning is a mathematical analysis framework that uses a combinatorial bound to describe the maximum tolerable error for a the output of a function \cite{valiant}.\\

\textbf{Definition 1}: \textit{In the case of machine learning, a concept is said to be PAC learnable if training results in a model that produces errors less than or equal to $(1 + \epsilon)$ times the optimum solution with a probability of at least ($1-\delta$).} \\

We extend this concept to robotic search by using it to place a combinatorial bound on the total distance traveled, a more relevant metric to real-time search than the length of the discovered shortest path.\\

\textbf{Definition 2}: \textit{An algorithm is PAC-admissible iff it produces errors less than or equal to $(1 + \epsilon)$ times the optimum solution with a probability of at least ($1-\delta$) where $0 < \delta < 1$}.

\section{PAC-Admissible Search -- $\delta \epsilon \alpha^*$} 
\subsection{The need for a combinatorial bound}
We propose $\delta \epsilon \alpha^*$, a PAC-admissible search algorithm that is fit for use in real-time, robotic search. The use of a combinatorial bound allows us to combine the $\alpha$-perimeters of AlphA* into a single $\alpha$-perimeter that no longer produces an unbounded amount of exploration or backtracking.
To illustrate this, first we must formally define an $\epsilon$-admissible algorithm:\\

\textbf{Definition 3:} \textit{An algorithm is said to be $\epsilon$-admissible iff its solutions are guaranteed to be no worse than $(1+\epsilon)$ times the optimal solution.}\\

From this definition, AlphA* proved that given a strictly admissible heuristic $h$, an algorithm is $\epsilon$-admissible when $1+\epsilon \geq \frac{1+\Lambda}{1+\lambda}, -1 < \lambda \leq \Lambda$. Their proof is as follows: \\

\textbf{Theorem 1.} \textit{If} $h$ \textit{is admissible, then AlphA* is $\epsilon$-admissible when}\\ 
$$1 + \epsilon \geq \frac{1+\Lambda}{1+\lambda},\; -1 < \lambda \leq \Lambda$$
\textit{Proof.} It must be shown that $C(t) \leq (1+\epsilon)C^*$, where $C(t)$ is the cost of the actual solution path and $C^*$ is the cost of the shortest possible path.
\begin{itemize}
\item[] $f(n) = \underset{n' \in \text{OPEN}}{\min}f(n’)$
\item[] Let $n_0$ be the node in the search frontier with the lowest $f$ value.
\item[] There is a least one node, $n^*_i$, from the optimal solution path in OPEN, because the search frontier is continuous.
\item[] Let $n^*_0$ be the shallowest $n^*_i$.
\item[] We know that $f(n_0^*) \leq C^*$, because $h$ is admissible.  
\item[] By definition, we have $f(n_0) \leq f(n^*_0)$
\item[] From $f_\alpha$ and $w_\alpha$ we have 
    \subitem $(1+\lambda)f(t) \leq (1+\Lambda)f(n_0) \implies f(t) \leq \frac{1+\Lambda}{1+\lambda} f(n^*_0), \; -1 < \lambda \leq \Lambda$
\item[] Combined, this gives us 
    \subitem $C(t) = f(t) \leq \frac{1+\Lambda}{1+\lambda} f(n_0) \leq \frac{1+\Lambda}{1+\lambda} f(n^*_0) \leq \frac{1+\Lambda}{1+\lambda} C^*(t) \leq (1+\epsilon)C^*$\qed
\end{itemize}

AlphA* reduces computational requirements by relaxing admissibility criteria, allowing promissory paths can be exploited as long as possible. Following from this notion, $\delta \epsilon \alpha^*$ further relaxes admissibility criteria such that a promissory path can sometimes be continued even if the next node is not $\epsilon$-admissible, minimizing the number of possible paths considered.\\

\textbf{Definition 4}: \textit{A search is considered real-time if precomputation of the search path is not possible. Alternatively, a real-time search requires that the entire path be calculated incrementally, one node at a time.} \\

The definition of a real-time search implies that each considered node must be traversed, and that backtracking attempts must revisit nodes rather than skip over them. From this, we have $$C(t) = \sum_{i=0}^{i=\mu} \sigma(n_i)f(n_i)$$ where $\mu$ is the number of nodes considered and $\sigma(n)$ is equal to the number of times a node is visited. \\

Both strict and $\epsilon$-admissible search algorithms consider many possible paths in effort to find an optimal solution. This results in poor real-time search performance, as both the number of nodes considered ($\mu$) and $\sigma(n)$ will be high. By further relaxing node admissibility criteria, we can minimize $\mu$ and $\sigma(n)$ and thus minimize the final distance traveled $C(t)$. \\

\subsection{Overview of $\delta \epsilon \alpha^*$}
When continuing upon a promissory path is no longer $\epsilon$-admissible, AlphA* will follow one of two behaviors (aggressive vs non-aggressive), that must be pre-selected prior to the search. Neither of these behaviors produce acceptable results in real-time search due to excessive backtracking or overly aggressive continuation of the promissory path, respectively. \\

To solve this, $\delta \epsilon \alpha^*$ allows for both behaviors to be used within a search. The probability at which the aggressive behavior will occur upon termination of a promissory path is described by $(1 - \delta)$ where $0 < \delta < 1$. A $\delta$ value of 1 would result in a real-time search that always uses the aggressive $\alpha$-perimeter, continuing upon the promissory path regardless of the $\epsilon$-admissibility of the next node. Similarly, a value of 0 would result in the search only exploring $\epsilon$-admissible nodes and a large amount of unnecessary backtracking. In other words -- due to the uncertainty associated with a real-time search, traversing some number non-$\epsilon$-admissible nodes results in a shorter, more direct path than backtracking upon termination of the admissible path, so long as there is a bound on the number of non-$\epsilon$-admissible nodes. \\

To illustrate how both behaviors can be used to set a combinatorial bound, consider the pseudocode for AlphA* upon termination of a promissory path in a real-time search:
\begin{itemize}
\item[] \texttt{if $f(n_i) > f(n_i^*)$:}
\subitem \texttt{if non-aggressive: $\alpha(n_i) := \alpha(n_0)$ }
\subitem \texttt{if aggressive: $\alpha(n_i) := \alpha(n')$ }
\item[] \texttt{else: $\alpha(n_i) := \alpha(n')$  // $n'$ is $\epsilon$-admissible so this is not aggressive.}
\end{itemize}
\pagebreak

Then, consider pseudocode for $\delta \epsilon \alpha^*$ upon termination of a promissory path in a real-time search:
\begin{itemize}
\item[] \texttt{if $f(n_i) > f(n_i^*)$:}
\subitem \texttt{if (1 - $\delta$)(100) < rand(0,100): $\alpha(n_i) := \alpha(n_0)$ }
\subitem \texttt{else: $\alpha(n_i) := \alpha(n')$ }
\item[] \texttt{else: $\alpha(n_i) := \alpha(n')$ }
\end{itemize}
The difference is subtle, yet significant. Recall that with AlphA*, only one $\alpha$-perimeter can be used per search. A similar effect could be achieved in $\delta \epsilon \alpha^*$ by setting $\delta = 1$ or $\delta = 0$, hence the requirement that $0 < \delta < 1$. \\

Also, recall that by the definition of a real-time search, nodes must be revisited if backtracking occurs, and that each node considered through exploration adds to the total distance traveled. We can prove that by minimizing backtracking and exploration, $\delta \epsilon \alpha^*$ will result a shorter distance traveled than an $\epsilon$-admissible real-time search. \\

\textbf{Theorem 2.} \textit{Given a real-time search that backtracks to a limited degree and a real-time search that backtracks to an unbounded degree, the cost of the bounded search $C_b(t)$ will be less than that of the unbounded search $C_u(t)$.}\\ 

\textbf{Theorem 3.} \textit{Given a real-time search that explores to a limited degree and a real-time search that explores to an unbounded degree, the cost of the bounded search $C_b(t)$ will be less than that of the unbounded search $C_u(t)$.}\\ 

We can prove theorems 2 and 3 in a single proof.\\

\textit{Proof.} It must be shown that $C_b(t) \leq C_u(t)$, where $C_b(t)$ is the cost of the bounded search and $C_u(t)$ is the cost of the  unbounded search.
\begin{itemize}
\item[] $C(t) = \sum_{i=0}^{i=\mu} \sigma(n_i)f(n_i)$, $C_b(t) = \sum_{i=0}^{i=\mu_b} \sigma_b(n_i)f(n_i)$, $C_u(t) = \sum_{i=0}^{i=\mu_u} \sigma_u(n_i)f(n_i)$
\item[] Let $\mu_b$ be the number of nodes considered by the search with bounded exploration. 
\item[] Let $\mu_u$ be the number of nodes considered by search with unbounded exploration. 

\item[] Let $\sigma_b(n_i)$ be the number of times a node visited by the search with bounded backtracking.
\item[] Let $\sigma_u(n_i)$ be the number of times a node visited by the search with bounded backtracking.
\subitem By definition, $\mu_b \leq \mu_u$ and $\sigma_b(n_i) \leq \sigma_u(n_i)$.
\item[] From this, we have 
\subitem $C_b(t) = \sum_{i=0}^{i=\mu_b} \sigma_b(n_i)f(n_i) \leq C_u(t) = \sum_{i=0}^{i=\mu_u} \sigma_u(n_i)f(n_i)\qed$
\end{itemize}

\section{Conclusion}
We have shown that it is possible to account for uncertainties in the a real-time robotic search environment by relaxing admissibility criteria through the use of a combinatorial bound. This method places additional weight on a promissory path while setting a probabilistic limit on the number of non-$\epsilon$-admissible nodes that are visited.

\bibliographystyle{asmems4}

\end{document}